%% file: main.tex
\documentclass{article}

    \usepackage[preprint]{neurips_2020}

\input{math_commands.tex}

\usepackage{hyperref}
\usepackage{url}
\usepackage{wrapfig}
\usepackage{makecell}
\usepackage{color}
\usepackage{graphicx}
\usepackage{amsmath}
\usepackage{amssymb}
\usepackage{mathrsfs}
\usepackage{tabularx}
\usepackage{multirow}
\usepackage{makecell}
\usepackage{enumitem}
\usepackage{booktabs}
\usepackage{minitoc}
\usepackage{etoc}
\usepackage{listings}
\usepackage{algorithm}
\usepackage{algorithmic}
\usepackage{setspace}
\usepackage{bbding}
\usepackage{natbib}
\setcitestyle{numbers,square}

\newtheorem{definition}{Definition}

\title{CausalTime: Realistically Generated Time-series for Benchmarking of Causal Discovery}

\author{Yuxiao Cheng\textsuperscript{1}\footnotemark[1]~~~~~~~~ Ziqian Wang\textsuperscript{1}\thanks{Equal Contribution}~~~~~~~~ Tingxiong Xiao\textsuperscript{1}~~~~~~~~ Qin Zhong\textsuperscript{3} \\
\textbf{Jinli Suo\textsuperscript{12}\thanks{Corresponding author}~~~~~~~~ Kunlun He\textsuperscript{3}\footnotemark[1]}\\
\textsuperscript{1}Department of Automation, Tsinghua University\\
\textsuperscript{2}Institute for Brain and Cognitive Science, Tsinghua University (THUIBCS)\\
\textsuperscript{3}Chinese PLA General Hospital\\
\texttt{{cyx22,ziqian-w20,xtx22}@mails.tsinghua.edu.cn} \\
\texttt{{zhongqin1,kunlunhe}@plagh.org}~~~~ \texttt{jlsuo@tsinghua.edu.cn} \\
}

\begin{document}

\maketitle

\begin{abstract}
Time-series causal discovery (TSCD) is a fundamental problem of machine learning. However,  existing synthetic datasets cannot properly evaluate or predict the algorithms' performance on real data. This study introduces the CausalTime pipeline to generate time-series that highly resemble the real data and with ground truth causal graphs for quantitative performance evaluation. The pipeline starts from real observations in a specific scenario and produces a matching benchmark dataset. Firstly, we harness deep neural networks along with normalizing flow to accurately capture realistic dynamics. Secondly, we extract hypothesized causal graphs by performing importance analysis on the neural network or leveraging prior knowledge. Thirdly, we derive the ground truth causal graphs by splitting the causal model into causal term, residual term, and noise term. Lastly, using the fitted network and the derived causal graph, we generate corresponding versatile time-series proper for algorithm assessment. In the experiments, we validate the fidelity of the generated data through qualitative and quantitative experiments, followed by a benchmarking of existing TSCD algorithms using these generated datasets. CausalTime offers a feasible solution to evaluating TSCD algorithms in real applications and can be generalized to a wide range of fields. For easy use of the proposed approach, we also provide a user-friendly website, hosted on \url{www.causaltime.cc}.
\end{abstract}

\section{Introduction}

Inferring causal structures from time-series, i.e., time-series causal discovery (TSCD), is a fundamental problem in machine learning. It goes beyond prediction or forecasting by revealing the complex interactions buried under multi-variate time-series. Recently, many algorithms have been proposed %in this area 
\citep{loweAmortizedCausalDiscovery2022, liCausalDiscoveryPhysical2020, wuGrangerCausalInference2022, brouwerLatentConvergentCross2021} and achieved satisfactory performance, i.e., the discovered causal graphs are close to the ground-truth counterparts. Under some settings, the causal discovery results are nearly perfect, with AUROC scores approaching 1.

However, the benchmarks for TSCD algorithms do not suffice for the performance evaluation. First of all, for the statistical significance of the quantitative evaluation results, the datasets need to be improved in terms of quality and quantity. Next, the current datasets are limited to several fields and do not cover wide application directions. More importantly, the datasets with ground-truth causal graphs %can be improved since the quality and quantity for baseline choices are limited, and 
are synthesized and might deviate from the true data-generating process, so the scores may not reflect the performance on real data \citep{reisachBewareSimulatedDAG2021}. 
%we ``realistic / authentic data'', ``ground truth graph'', and ``generalizable to other fields'' at the same time.

Despite the fact that recent works also propose better benchmarks for time-series causal discovery \citep{lawrenceDataGeneratingProcess2021, rungeCausalityClimateCompetition2020}, as well as static settings \citep{goblerTextttCausalAssemblyGenerating2023, chevalleyCausalBenchLargescaleBenchmark2023, chevalleyCausalBenchChallengeMachine2023}. Current TSCD algorithms often incorporate three types of datasets: % {numerical datasets}, {quasi-real datasets}, and {real datasets}. 
{\em Numerical datasets}, e.g., VAR (vector auto-regression) and Lorenz-96 \citep{karimiExtensiveChaosLorenz962010}, are simulated using closed-form equations. Although some of these equations (Lorenz-96) are inspired by real application scenarios, e.g., climate dynamics, they are over-simplified and have very limited generalizability to real-world applications \citep{rungeCausalityClimateCompetition2020}. {\em Quasi-real datasets} are composed of time-series generated with manually designed dynamics that mimic real counterparts under a certain scenario. For example, DREAM3 \citep{prillRigorousAssessmentSystems2010} is a dataset simulated using gene expression and regulation dynamics, and NetSim \citep{smithNetworkModellingMethods2011} is generated by simulating interactions between human brain regions under observation of fMRI. The problem with this type of dataset is that it only covers a few research areas with underlying mechanisms relatively clearly known. For fields such as healthcare or finance, it is hard or even impossible to generate realistic time-series with manually designed dynamics. {\em Real datasets} (such as MoCap \citep{tankNeuralGrangerCausality2022}, S\&P 100 stock returns \citep{pamfilDYNOTEARSStructureLearning2020}) do not have the above-mentioned problem, but the dealbreaker is that the ground truth causal graph is mostly inaccessible, and we have to resort to some ad hoc explanations. % because we can never tell if a discovered causal relationship is ``correct'' in many complex scenarios.
As shown in Table \ref{tab_tradeoff}, currently available benchmarking tools cannot support a comprehensive evaluation of the time-series causal discovery algorithm. Therefore, an approach for generating benchmarks that highly mimic the real data in different scenarios and with true causal graphs is highly demanded. 
%Consequently, to scientifically evaluate the performance of the time-series causal discovery algorithm, a tradeoff must be made. 

\begin{table}[ht] % {l}{8cm}
\centering
\small
% \vspace{-15pt}
\caption{Comparison of benchmarks for time-series causal discovery evaluation.}
\label{tab_tradeoff}
% \vspace{3pt}
\setlength\tabcolsep{6pt}
\renewcommand{\arraystretch}{1.05}
\begin{tabular}{c|cccc}
 \hline
{~~Datasets~~}                                    & {~~Numerical~~}     & {~~Quasi-real~~}  & {~~Real~~}    & \textbf{~~CausalTime (Ours)~~} \\  \hline
{Realistic Data}                                  & Low                 & Moderate          & Very High     & High         \\
{\makecell[c]{With True Causal Graph}}            & \Checkmark          & \Checkmark        & \XSolidBrush  & \Checkmark         \\
{\makecell[c]{Generalizable to Diverse Fields}}   & \XSolidBrush        & \XSolidBrush      & \Checkmark    & \Checkmark         \\ \hline
\end{tabular}
\end{table}

In this work, we propose a novel pipeline capable of generating realistic time-series along with a ground truth causal graph and is generalizable to different fields, named \textbf{CausalTime}. 
The process of generating time-series with a given causal graph can be implemented using the autoregression model, however, pursuing a causal graph that matches the target time-series with high accuracy is nontrivial, especially for the data with little prior knowledge about the underlying causal mechanism. 
To address this issue, we propose to use a deep neural network to fit the observed data with high accuracy, and then retrieve a causal graph from the network or from prior knowledge that holds high data fidelity. Specifically, we first obtain a hypothesized causal graph by performing importance analysis on the neural network or leveraging prior knowledge, and then split the functional causal model into causal term, residual term, and noise term. The split model can naturally generate time-series matching the original data observations well. It is worth noting that the retrieval of the causal graph is not a causal discovery process and does not necessarily uncover the underlying causal relationship, but can produce realistic time-series to serve as the benchmark of causal discovery algorithms. 

%performing causal discovery in the process of generating a causal discovery dataset, we extract As a result, a new time-series with ground truth causal graphs can be generated using the autoregression model. 
%Our aim is to first and then generate new time-series with a ground truth causal graph.
%A plausible approach would be to first extract a causal graph with existing TSCD algorithms and then use the causal graph to fit the series. However, extracting reliable causal graph from real time-series from is hard, and the accuracy of the following fitting process with inaccurate causal graph may be hampered. 
%More importantly, this kind of pipeline seems circular and unnatural.

Our benchmark is open-source and user-friendly, we host our website at \url{www.causaltime.cc}. Specifically, our contributions include:
\begin{itemize}[leftmargin=*]
\setlength{\itemsep}{0pt}
\setlength{\parsep}{0pt}
\setlength{\parskip}{0pt}
    \item We propose CausalTime, a pipeline to generate realistic time-series with ground truth causal graphs, which can be applied to diverse fields and provide new choices for evaluating TSCD algorithms.
    \item We perform qualitative and quantitative experiments to validate that the generated time-series preserves the characteristics of the original time-series. 
    \item We evaluate several existing TSCD algorithms on the generated datasets, providing some guidelines for algorithm comparison, choice, as well as improvement.
\end{itemize}

\section{Related Works}
\textbf{Causal Discovery.~~~~} Causal Discovery (or Causal Structural Learning), including static settings and dynamic time-series, has been a hot topic in machine learning and made big progress in the past decades. The methods can be roughly categorized into multiple classes. (i) \textit{Constraint-based approaches}, such as PC \citep{spirtesAlgorithmFastRecovery1991}, FCI \citep{spirtesCausationPredictionSearch2000}, and PCMCI \citep{rungeDetectingQuantifyingCausal2019, rungeDiscoveringContemporaneousLagged2020, gerhardusHighrecallCausalDiscovery2020}, build causal graphs by performing conditional independence tests. (ii) \textit{Score-based learning algorithms} which include penalized Neural Ordinary Differential Equations and acyclicity constraint \citep{bellotNeuralGraphicalModelling2022} \citep{pamfilDYNOTEARSStructureLearning2020}. (iii) Approaches based on \textit{Additive Noise Model (ANM)} that infer causal graph based on additive noise assumption \citep{shimizuLinearNonGaussianAcyclic2006, hoyerNonlinearCausalDiscovery2008}. ANM is extended by \citet{hoyerNonlinearCausalDiscovery2008} to nonlinear models with almost any nonlinearities. (iv) \textit{Granger-causality-based} approaches. Granger causality was initially introduced by \citet{grangerInvestigatingCausalRelations1969} who proposed to analyze the temporal causal relationships by testing the help of a time-series on predicting another time-series. Recently, Deep Neural Networks (NNs) have been widely applied to nonlinear Granger causality discovery. \citep{wuGrangerCausalInference2022, tankNeuralGrangerCausality2022, khannaEconomyStatisticalRecurrent2020, loweAmortizedCausalDiscovery2022, chengCUTSNeuralCausal2023}. (v) \textit{Convergent Cross Mapping (CCM)} proposed by \citet{sugiharaDetectingCausalityComplex2012} that reconstructs nonlinear state space for nonseparable weakly connected dynamic systems. This approach is later extended to situations of synchrony, confounding, or sporadic time-series \citep{yeDistinguishingTimedelayedCausal2015, benkoCompleteInferenceCausal2020, brouwerLatentConvergentCross2021}. 
% \cyx{More discussion for each category.} 
The rich literature in this direction requires effective quantitative evaluation and progress in this direction also inspires designing new benchmarking methods. In this paper, we propose to generate benchmark datasets using causal models.

\textbf{Benchmarks for Causal Discovery.~~~~}  Benchmarking is of crucial importance for algorithm design and applications. Researchers have proposed different datasets and evaluation metrics %In this work, we mainly focus on time-series causal discovery (TSCD). However, we provide related works 
for causal discovery under both static and time-series settings.
(i) {\textit{Static settings.~~~~}} Numerical, quai-real, and real datasets are all widely used in static causal discovery. Numerical datasets include datasets simulated using linear, polynomial, or triangular functions \citep{hoyerNonlinearCausalDiscovery2008, mooijCausalDiscoveryCyclic2011, spirtesAlgorithmFastRecovery1991, zhengDAGsNOTEARS2018}; Quai-real datasets are generated under physical laws (e.g. double pendulum \citep{brouwerLatentConvergentCross2021}) or realistic scenarios (e.g. synthetic twin birth datasets \citep{geffnerDeepEndtoendCausal2022}, alarm message system for patient monitoring \citep{scutariLearningBayesianNetworks2010, lippeEfficientNeuralCausal2021}, neural activity data \citep{brouwerLatentConvergentCross2021}, and gene expression data \citep{vandenbulckeSynTReNGeneratorSynthetic2006}); Real datasets are less frequently used. Examples include ``Old Faithful'' dataset on volcano eruptions \citep{hoyerNonlinearCausalDiscovery2008}, and expression levels of proteins and phospholipids in human immune system cell \citep{zhengDAGsNOTEARS2018}. Recently, \citet{goblerTextttCausalAssemblyGenerating2023} proposes a novel pipeline, i.e., causalAssembly, generating realistic and complex assembly lines in a manufacturing scenario. \citet{chevalleyCausalBenchLargescaleBenchmark2023} and \citet{chevalleyCausalBenchChallengeMachine2023} on the other hand, provides CausalBench, a set of benchmarks on real data from large-scale single-cell perturbation. Although causalAssembly and CausalBench are carefully designed, they are restricted in certain research fields where the dynamics can be easily replicated or the ground truth causal relationships can be acquired by performing interventions.
(ii) {\textit{Time-series settings.~~~~}} In time-series settings, widely used numerical datasets include VAR and Lorenz-96 \citep{tankNeuralGrangerCausality2022, chengCUTSNeuralCausal2023, khannaEconomyStatisticalRecurrent2020, bellotNeuralGraphicalModelling2022}; quasi-real datasets include NetSim \citep{loweAmortizedCausalDiscovery2022}, Dream-3 / Dream-4 \citep{tankNeuralGrangerCausality2022}, and finance dataset simulated using Fama-French Three-Factor Model \citep{nautaCausalDiscoveryAttentionbased2019}; real datasets include MoCap dataset for human motion data \citep{tankNeuralGrangerCausality2022}, S\&P 100 stock data \citep{pamfilDYNOTEARSStructureLearning2020}, tropical climate data \citep{rungeDetectingQuantifyingCausal2019}, and complex ecosystem data \citep{sugiharaDetectingCausalityComplex2012}. Other than these datasets, there are several works providing novel benchmarks with ground-truth causal graphs. CauseMe \citep{rungeCausalityClimateCompetition2020, rungeInferringCausationTime2019} provides a platform\footnote{causeme.net} for numerical, quasi-real, as well as real datasets, which are mainly based on TSCD challenges on climate scenarios. However, although the platform is well-designed and user-friendly, it did not alleviate the tradeoff among fidelity, ground truth availability, and domain generalizability. For example, the numerical datasets in CauseMe are still not realistic, and the ground truth causal graphs for real datasets are still based on domain prior knowledge that may not be correct. \citep{lawrenceDataGeneratingProcess2021} focuses on generating time-series datasets that go beyond CauseMe. Their framework allows researchers to generate numerous data with various properties flexibly. The ground-truth graphs for their generated datasets are exact, but the functional dependencies in \citet{lawrenceDataGeneratingProcess2021} are still manually designed and may not reflect real dynamics in natural scenarios. As a result, their generated datasets are still categorized into numerical datasets in table \ref{tab_tradeoff}, although with far higher flexibility.

Recently, neural networks have been extensively studied for their capability of generating time-series \citet{yoonTimeseriesGenerativeAdversarial2019, jarrettTimeseriesGenerationContrastive2021, peiGeneratingRealWorldTime2021, kangGRATISGeneRAtingTIme2020, zhangGenerativeAdversarialNetwork2018, estebanRealvaluedMedicalTime2017}. However, the time-series generated with these methods are improper for benchmarking causal discovery, since causal graphs are not generated alongside the series. Therefore, we propose a pipeline to generate realistic time-series along with the ground truth causal graphs.

% \section{Preliminaries}

\section{The Proposed Time-series Generation Pipeline}
\label{method}
\begin{figure}[t]
    \centering
    \includegraphics[width=\linewidth]{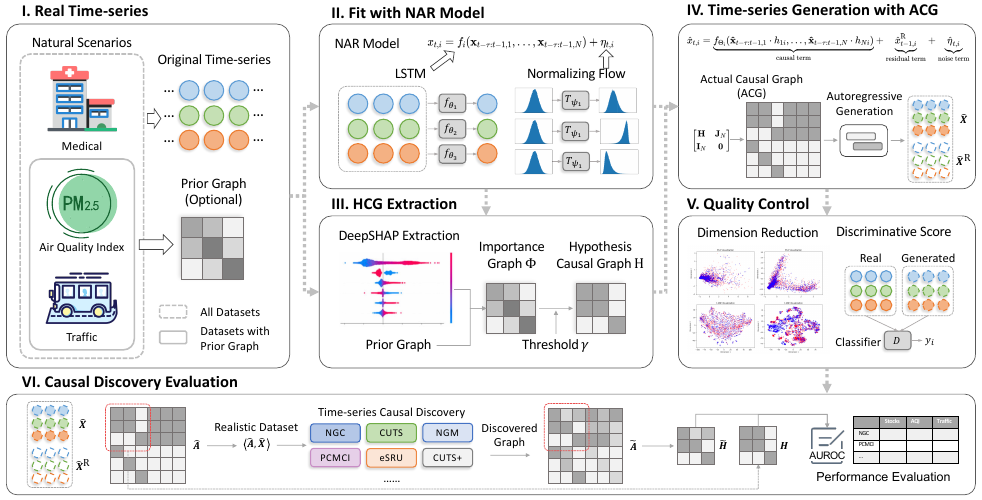}
    \vspace{-5mm}
    \caption{Illustration of our realistic time-series generation pipeline, which takes observations from real scenarios as input and produces benchmark data for performance evaluation of TSCD algorithms. The whole pipeline consists of several key steps, which are explained in Section \ref{method} in detail.}
    \label{tab_arch}
\end{figure}

Aiming at generating time-series that highly resemble observations from real scenarios, and with ground truth causal graphs, we propose a general framework to generate a causal graph built from the real observations and generate counterpart time-series that highly resemble these observations. The built causal graph serves as the ground truth lying under the generated counterpart, and they together serve as a benchmark for the TSCD algorithms. The whole pipeline consists of several key steps, as illustrated in Fig.~\ref{tab_arch}.  
%To generate realistic time-series, six steps are included in our pipeline. (i) \textit{Time-series Acquisition.} Collect real time-series from diverse fields. (ii) \textit{Time-series Fitting.} Fit the dynamic process of multivariate time-series with deep neural network and normalizing flow. (iii) \textit{Hypothetical Causal Graph Extraction.} Select the most contributing causal parents of each variable by applying Shapley values analysis or prior knowledge. (iv) \textit{Time-series Generation with Actual Causal Graph.} Generate new time-series with actual causal graph (ACG), where the NAR model is split into causal term, residual term, and noise term. As a result, the high accuracy of the fitting model is preserved, and the ground truth causal graph is acquired. (v) \textit{Time-series Quality Control.} Examine the fidelity of the generated time-series by calculating discriminative scores, and performing t-SNE \citep{maatenVisualizingDataUsing2008} and PCA \citep{bryantPrincipalcomponentsAnalysisExploratory1995} for visualized examination. and finally (vi) \textit{TSCD evaluation.} 

We would like to clarify that, our generation pipeline is based on several assumptions that are common in causal discovery literature: markovian condition, faithfulness, sufficiency, no instantaneous effect, and stationarity. We place the detailed discussion of these assumptions in Supplementary Section \ref{app_assumption} due to page limits.

\subsection{Causal Model}

Causal models in time-series are frequently represented as graphical models \citep{vowelsYaDagsSurvey2021, spirtesCausationPredictionSearch2000}. However, different from the classic Pearl's causality \citep{pearlCausalityModelsReasoning2009}, spatio-temporal structural dependency must be taken into account for time-series data. We denote a uniformly sampled observation of a dynamic system as $\mX = \{\vx_{1:T,i}\}_{i=1}^{N}$, in which $\vx_t$ is the sample vector at time point $t$ and consists of $N$ variables $\{x_{t,i}\}$, with $t\in \left\{ 1, ..., T \right\}$ and $i\in \left\{ 1, ..., N \right\}$. The structural causal model (SCM) for time-series \citep{rungeDetectingQuantifyingCausal2019} is 
$x_{t,i}=f_i \left(\mathcal{P}(x_{t,i}), \eta_{t,i}\right), ~~~ i=1,2,..., N$,
where $f_i$ is any (potentially) nonlinear function, $\eta_{t,i}$ denotes dynamic noise with mutual independence, and $\mathcal{P}(x_t^j)$ denotes the causal parents of $x_t^j$. This model is generalizable to most scenarios, but may bring obstacles for our implementation. In this paper, we consider the nonlinear autoregressive model (NAR), a slightly restricted class of SCM. 

\textbf{Nonlinear Autoregressive Model.}
We adopt the representation in many time-series causal discovery algorithms (\citep{tankNeuralGrangerCausality2022, loweAmortizedCausalDiscovery2022, chengCUTSNeuralCausal2023}), as well as \citet{lawrenceDataGeneratingProcess2021}'s time-series generation pipeline.  In a Nonlinear Autoregressive Model (NAR), the noise is assumed to be independent and additive, and each sampled variable $x_{t,i}$ is generated by the following equation:
\begin{equation}
\label{nar}
    x_{t,i}=f_i \left(\mathcal{P}(x_{t,i})\right) + \eta_{t,i}, ~~~ i=1,2,..., N.
\end{equation}
where $\mathcal{P}(\cdot)$ denotes parents in causal graph. We further assume that the maximal time lags for causal effects are limited. Then the model can be denoted as
$x_{t,i}=f_i \left(\{x_{\tau,j}\}_{x_{\tau,j}\in \mathcal{P}(x_{t,i})}\right) + \eta_{t,i}$.
Here $t-\tau\leq \tau_{\text{max}}, \forall x_{\tau,j}\in \mathcal{P}(x_{t,i})$, and $\tau_{\text{max}}$ denotes the maximal time lag. In causal discovery, time-homogeneity \citep{gongCausalDiscoveryTemporal2023} is often assumed, i.e., function $f_i$ and causal parents $\mathcal{P}$ is irrelevant to time. By summarizing temporal dependencies, the summary graph for causal models can be denoted with binary matrix $\mathbf{A}$, where its element 
$a_{ji}=\begin{cases}
1, & \exists \tau, s.t., x_{\tau,j}\in \mathcal{P}(x_{t,i}) \\
0, & \text{otherwise}
\end{cases}$. The dataset pair for causal discovery is $\left< \mathbf{X}, \mathbf{A} \right>$. TSCD targets to recover matrix $\mathbf{{A}}$ given time-series $\mathbf{X}$. However, since for most real time-series $\mathbf{X}$, causal graph $\mathbf{A}$ is unknown, benchmarking causal discovery algorithms with real time-series is generally inappropriate.

% In this work, we assume Causal Sufficiency, i.e., all common causes are included in the dataset and assume no instantaneous effects, i.e., $t-\tau \geq 1, \forall x_{\tau,j}\in \mathcal{P}(x_{t,i})$, resulting in an acyclic full time causal graph (although summary graph $\mathbf{A}$ may contain cycles). The detailed discussions of the assumptions are in Supplementary Section \ref{app_assumption}.

\subsection{Time-series Fitting}
\label{fitting}

After collecting real-time-series from diverse fields, we fit the dynamic process of multivariate time-series with a deep neural network and normalizing flow. 

\textbf{Time-series Fitting with Causally Disentangled Neural Network (CDNN).~~~~} To fit the observed time-series with a deep neural network and introduce casual graphs into the network's prediction of output series, %gain insight into the inner workings of neural networks and include causal graphs in the prediction model, 
\citet{tankNeuralGrangerCausality2022, khannaEconomyStatisticalRecurrent2020, chengCUTSNeuralCausal2023} separate the causal effects from the parents to each of individual output series using $N$ separate MLPs / LSTMs, which is referred to as component-wise MLP / LSTM (cMLP / cLSTM). In this paper, we follow \citep{chengCUTSHighdimensionalCausal2023}'s definition and refer to the component-wise neural networks as ``causally disentangled neural networks''.
\begin{definition}
\label{cdnn}
    Let $\mathbf{X} \in \mathbf{\mathcal{X}} \subset \mathbb{R}^{T\times N}, \mathbf{A} \in \mathbf{\mathcal{A}} \subset \left\{0,1\right\}^{n\times n}$ and $\mathbf{y} \in \mathbf{\mathcal{Y}} \subset \mathbb{R}^n$ be the input and output spaces. We say a neural network $\mathbf{f}_\Theta: \left<\mathbf{\mathcal{X}}, \mathbf{\mathcal{A}} \right> \rightarrow \mathbf{\mathcal{Y}}$ is a \textbf{causally disentangled neural network (CDNN)} if it has the form 
    \begin{equation}
        \mathbf{f}_\Theta(\mathbf{X}, \mathbf{A}) = \left[ f_{\Theta_1}(\mathbf{X} \odot \mathbf{a}_{:,1}), ..., f_{\Theta_n}(\mathbf{X} \odot \mathbf{a}_{:,n}) \right]^T,
    \end{equation}
    where $\mathbf{a}_{:,j}$ is the column vector of input causal adjacency matrix $\mathbf{A}$; $f_{\phi_j}: \mathbf{\mathcal{X}}_j \rightarrow \mathbf{\mathcal{Y}}_j$, with $\mathbf{\mathcal{X}}_j \subset \mathbb{R}^{n}$ and $\mathbf{\mathcal{Y}}_j \subset \mathbb{R}$; the operator $\odot$ is defined as $f_{\phi_j}(\mathbf{X} \odot \mathbf{a}_{:,j}) \overset{\Delta}{=} f_{\phi_j}\left(\left\{ \mathbf{x}_1 \cdot a_{1j}, ..., \mathbf{x}_N \cdot a_{Nj} \right\}\right)$.
\end{definition}

So far function $f_{\Theta_j}(\cdot)$ acts as the neural network function used to approximate $f_j(\cdot)$ in Equation (\ref{nar}). Since we assume no prior on the underlying causal relationships, to extract the dynamics of the time-series with high accuracy, we fit the generation process with all historical variables (with maximal time lag $\tau_{\text{max}}$, which is discussed in Supplementary Section \ref{app_assumption}) and obtain %. As a result, the causal graph in the fitting model is 
a fully connected graph. Specifically, we assume that
\begin{equation}
    x_{t,i}=f_i \left(\mathbf{x}_{t-\tau:t-1,1}, ..., \mathbf{x}_{t-\tau:t-1,N} \right) + \eta_{t,i}, ~~~ i=1,2,..., N.
\end{equation}

In the following, we omit the time dimension of $\mathbf{x}_{t-\tau:t-1,j}$ and denote it with $\mathbf{x}_j$. Using a CDNN in Definition \ref{cdnn}, we can approximate $f_i(\cdot)$ with $f_{\Theta_i}(\cdot)$. \citep{tankNeuralGrangerCausality2022} and \citep{chengCUTSNeuralCausal2023} implement CDNN with component-wise MLP / LSTM (cMLP / cLSTM), but the structure is highly redundant because it consists of $N$ distinct neural networks. 

\textbf{Implementation of CDNN.~~~~} The implementation of CDNN can vary. For example, \citet{chengCUTSHighdimensionalCausal2023} explores enhancing causal discovery with a message-passing-based neural network, which is a special version of CDNN with extensive weight sharing. In this work, we utilize an LSTM-based CDNN with a shared decoder (with implementation details shown in Supplementary Section \ref{app_details}). Moreover, we perform scheduled sampling \citep{bengioScheduledSamplingSequence2015} to alleviate the accumulated error when performing autoregressive generation.

\textbf{Noise Distribution Fitting by Normalizing Flow.~~~~} After approximating the functional term $f_i(\cdot)$ with $f_{\Theta_i}(\cdot)$, we then approximate noise term $\eta_{t,i}$ with Normalizing Flow (NF) \citep{kobyzevNormalizingFlowsIntroduction2021, papamakariosNormalizingFlowsProbabilistic2021}. The main process is described as
\begin{equation}
    \hat{\eta}_{t,i} = T_{\psi_i}(u), \text{ where } u \sim p_u(u),
\end{equation}
in which $T_{\psi_i}(\cdot)$ is an invertible and differentiable transformation implemented with neural network, $p_u(u)$ is the base distribution (normal distribution in our pipeline), and $p_{\hat{\eta}_i}(\hat{\eta}_{t,i})=p_u(T^{-1}_{\psi_i}(\hat{\eta}_{t,i})) \frac{\partial}{\partial \hat{\eta}_{t,i}} T^{-1}_{\psi_i}(\hat{\eta}_{t,i})$. Then, the optimization problem can be formulated as $\max_{\psi_i} \sum_{t=1}^N \log p_u(T^{-1}_{\psi_i}(\eta_{t,i})) + \log \frac{\partial}{\partial \eta_{t,i}} T^{-1}_{\psi_i}(\eta_{t,i}) $

\subsection{Extraction of Hypothetical Causal Graph}
\label{hcg}

%In the aforementioned sections, we fit the time-series with a fully connected graph. As a result, 
In the fully connected graph, all $N$ variables contribute to each prediction, which fits the observations quite well but is over-complicated than the latent causal graph. We now proceed to extract a hypothetical causal graph (HCG) $\mathbf{H}$ by identifying the most contributing variables in the prediction model $f_{\Theta_i}(\cdot)$. We would like to clarify that, extracting HCG is \textbf{not} causal discovery, and it instead targets % The goal of this step is 
to identify the contributing causal parents while preserving the fidelity of the fitting model. Two options are included in our pipeline: i) HCG extraction with DeepSHAP; ii) HCG extraction with prior knowledge.

\textbf{HCG Extraction with DeepSHAP.~~~~} Shapley values \citep{sundararajanManyShapleyValues2020} are frequently used to assign feature importance for regression models. It originates from cooperative game theory \citep{kuhnContributionsTheoryGames2016}, and has recently been developed to interpret deep learning models (DeepSHAP) \citep{lundbergUnifiedApproachInterpreting2017, chenExplainingSeriesModels2022}. For each prediction model $f_{\Theta_i}(\cdot)$, the calculated importance of each input time-series $\mathbf{x}_{t-\tau:t-1, j}$ by DeepSHAP is $\phi_{ji}$. By assigning importance values from each time-series $j$ to $i$, we get the importance matrix $\Phi$. After we set the sparsity $\sigma$ of a HCG, a threshold $\gamma$ can be calculated with cumulative distribution, i.e., $\gamma = F_\phi^{-1}(\sigma)$, where $F_\phi(\cdot)$ is the cumulative distribution of $\phi_{ji}$ if we assume all $\phi_{ji}$ are i.i.d.

\textbf{HCG Extraction with Prior Knowledge.~~~~} Time-series in some fields, e.g., weather or traffic, relationships between each variable are highly relevant to geometry distances. For example, air quality or traffic flows in a certain area can largely affect that in a nearby area. As a result, geometry graphs can serve as hypothetical causal graphs (HCGs) in these fields, we show the HCG calculation process in Supplementary Section \ref{app_details} for this case.

The extracted HCGs is not the ground truth causal graph of time-series $\mathbf{X}$, because time-series $\mathbf{X}$ is not generated by the corresponding NAR or SCM model. A trivial solution would be running auto-regressive generation of time-series by setting input of non-causal terms to zero, i.e., 
\begin{equation}
    \hat{x}_{t,i} = f_{\Theta_i} \left(\mathbf{\hat{x}}_{t-\tau:t-1,1}\cdot h_{1i}, ..., \mathbf{\hat{x}}_{t-\tau:t-1,N}\cdot h_{Ni} \right) + \hat{\eta}_{t,i}, ~~~ i=1,2,..., N.\label{hcg}
\end{equation}
with $\{h_{ji}\}$ being the entries in the HCG $\mathbf{H}$. The fidelity of the fitting model $f_{\Theta_i}(\cdot)$ in Eq.~\ref{hcg} is greatly hampered by only including a subset of the variables. In the following, we introduce another way to generate a time-series with an actual causal graph, and most importantly, without losing fidelity.

\subsection{Time-series Generation from the Actual Causal Graph}
\label{acg}

%In the following, we introduce another way to generate time-series with an actual causal graph, and most importantly, without losing fidelity.
To acquire the Actual Causal Graphs (ACGs) with high data fidelity, we propose to split the NAR model taking the form of a fully connected graph into causal term, residual term, and noise term, i.e.,
\begin{equation}
\label{eqn_split}
    \hat{x}_{t,i} = \underbrace{f_{\Theta_i} \left(\mathbf{\hat{x}}_{t-\tau:t-1,1}\cdot h_{1i}, ..., \mathbf{\hat{x}}_{t-\tau:t-1,N}\cdot h_{Ni} \right)}_{\text{causal term}} +  \underbrace{\hat{x}^{\text{R}}_{t-1,i}}_{\text{residual term}} + \underbrace{\hat{\eta}_{t,i}}_{\text{noise term}},
\end{equation}
Where the residual term $\hat{x}^{\text{R}}_{t-1,i}$ indicates the ``causal effect'' of non-parent time-series of time-series $i$ in HCG $\mathbf{H}$. In other words, causal terms represent the \textbf{``major parts''} of causal effects, and the residual term represents the remaining parts. Mathematically, $\hat{x}^{\text{R}}_{t-1,i}$ is calculated as 
\begin{equation}
\label{eqn_res}
    \hat{x}^{\text{R}}_{t-1,i} = f_{\Theta_i} \left(\mathbf{\hat{x}}_{t-\tau:t-1,1}, ..., \mathbf{\hat{x}}_{t-\tau:t-1,N} \right)  - f_{\Theta_i} \left(\mathbf{\hat{x}}_{t-\tau:t-1,1}\cdot h_{1i}, ..., \mathbf{\hat{x}}_{t-\tau:t-1,N}\cdot h_{Ni} \right).
\end{equation}

When treated as a generation model, $\mathbf{x}_j \rightarrow \mathbf{x}_i^{\text{R}}$ contains instantaneous effects, however, which does not affect the causal discovery result of $\mathbf{x}_j \rightarrow \mathbf{x}_i$ for most existing TSCD approaches, as we discussed in Supplementary Section \ref{app_inst}. 
After randomly selecting an initial sequence $\mathbf{X}_{t_0-\tau_{\text{max}}:t_0-1}$ from the original time-series $\mathbf{X}$, $\mathbf{x}_i$ and $\mathbf{x}_i^{\text{R}}$ is generated via the auto-regressive model, i.e., the prediction results from the previous time step are used for generating the following time step. Our final generated time-series include all $\mathbf{x}_i$ and $\mathbf{x}_i^{\text{R}}$, i.e., a total of $2N$ time-series are generated, and the ACG $\mathbf{\hat{A}}$ is of size $2N\times 2N$:
\begin{equation}
\label{eqn_matrix}
\mathbf{\hat{X}} = \left\{ \mathbf{\hat{X}}_{t_0:t_0+T}; \mathbf{\hat{X}}^{\text{R}}_{t_0:t_0+T}  \right\}, \ \ 
 \mathbf{\hat{A}} = \left[
	               \begin{array}{c c}
	               \mathbf{{H}}    & \mathbf{J}_N \\
	               \mathbf{I}_N        & \mathbf{0}
	               \end{array}
	        \right]_{2N\times 2N}
\end{equation}
where $\mathbf{J}_N$ is an all-one matrix, $T$ is the total length of the generated time-series. With the prediction model $f_{\theta_i}$, normalizing flow model $T_{\psi_i}$, and ACG $\mathbf{A}$, we can obtain the final dataset $\left< \mathbf{\hat{X}}, \mathbf{\hat{A}} \right>$. $\mathbf{\hat{X}}$ is then feeded to TSCD algorithms to recover matrix $\mathbf{{A}}$ given time-series $\mathbf{\hat{X}}$.
% \textbf{Cross-correlations Score.} \cyx{TO-DO}

\section{Experiments}
In this section, we demonstrate the CausalTime dataset built with the proposed pipeline,  visualize and quantify the fidelity of the generated time-series, and then benchmark the performance of existing TSCD algorithms on CausalTime. %For statistical evaluation of the quantitative, we averaged over results on simulations from 5 different random seeds. In the following experiments, we also show the standard derivations.

\subsection{Statistics of the Benchmark Datasets}

Theoretically, the proposed pipeline is generalizable to diverse fields. Here we generate 3 types of benchmark time-series from weather, traffic, and healthcare scenarios respectively, as illustrated in Fig.~\ref{fig_data}. As for the time-series of weather and traffic, relationships between two variables are highly relevant to their geometric distances, i.e., there exists a prior graph, while there is no such prior in the healthcare series. The detailed descriptions of three benchmark subsets are as follows:

% \begin{wrapfigure}{r}{8cm}
\begin{figure}[ht]
    \centering
    \vspace{-1mm}
    \includegraphics[width=0.94\linewidth]{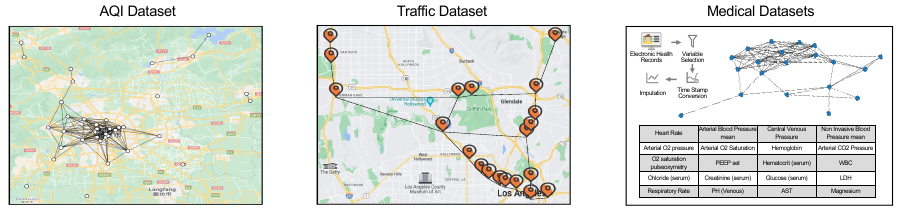}
    \vspace{-3mm}
    \caption{Visualization of three subsets in CausalTime. For AQI and Traffic, we overlay the ground truth causal graphs onto the map.}
    \vspace{-1mm}
    \label{fig_data}
\end{figure}
% \end{wrapfigure}

\begin{enumerate}[leftmargin=*]
\setlength{\itemsep}{0pt}
\setlength{\parsep}{0pt}
\setlength{\parskip}{0pt}
    \item \textbf{Air Quality Index (AQI)} is a subset of several air quality features from 36 monitoring stations spread across Chinese cities\footnote{https://www.microsoft.com/en-us/research/project/urban-computing/}, with an hourly measurement over one year. We consider the PM2.5 pollution index in the dataset. The total length of the dataset is L = 8760 and the number of nodes is N = 36. We acquire the prior graph by computing the pairwise distances between sensors (Supplementary Section \ref{app_details}).
    \item \textbf{Traffic} subset is built from the time-series collected by traffic sensors in the San Francisco Bay Area\footnote{https://pems.dot.ca.gov/}. The total length of the dataset is L = 52116 and we include 20 nodes, i.e., $N = 20$. The prior graph is also calculated with the geographical distance (Supplementary Section \ref{app_details}). 
    \item \textbf{Medical} subset is from MIMIC-4, which is a database that provides critical care data for over 40,000 patients admitted to intensive care units \citep{johnsonMIMICIVFreelyAccessible2023}. We select 20 most frequently tested vital signs and ``chartevents'' from 1000 patients, which are then transformed into time-series where each time point represents a 2-hour interval. The missing entries are imputed using the nearest interpolation. For this dataset, a prior graph is unavailable because of the extremely complex dynamics.
    % \item {\color{red}\textbf{Stocks} subset is a financial dataset of daily stock returns of companies in China, acquired with eFinance package\footnote{https://github.com/Micro-sheep/efinance}. We include 100 representative stocks from the areas of consumer, energy, financial, healthcare, industrial, technology, and utilities, respectively. The total length of the dataset is L = XXX and the number of nodes is N = XXX. The prior graph is not available in this case. }%available because the relationships between stocks are extremely complex and unknown to researchers.
\end{enumerate}

\subsection{fidelity of the Generated Time-series}

To qualitatively and quantitatively analyze the fidelity of the generated time-series, we utilize PCA \citet{bryantPrincipalcomponentsAnalysisExploratory1995} and t-SNE \citet{maatenVisualizingDataUsing2008} dimension reduction visualization, neural-network-based discriminative score, and MMD score of real and synthetic feature vectors to evaluate whether our generated time-series is realistic.

\begin{figure}[ht]
    \centering
    \includegraphics[width=\linewidth]{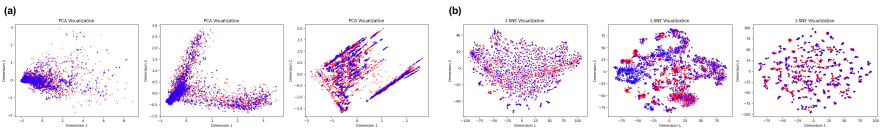}
    \vspace{-5mm}
    \caption{Visualization of the similarity between generated and original time-series in a low dimensional (2D) space, where original and generated series are shown in blue and red, respectively. (a) and (b) plots the two components of PCA and t-SNE on three datasets.}
    \vspace{-1mm}
    \label{fig_dim_rd}
\end{figure}

\textbf{Visualization via Dimension Reduction.~~~~} 
To judge the fidelity of the generated time-series, we project the time-series features to a two-dimensional space, and assess their similarity by comparing the dimension reduction results. % and visually comparing if the two distributions overlap.
%Both linear (PCA) and nonlinear (t-SNE) approaches are used to ensure better visualization. 
After splitting the original and generated time-series into short sequences (length of 5), we perform dimension reduction via linear (PCA) and nonlinear (t-SNE) approaches on three generated datasets and visualize the difference explicitly, as shown in Figure \ref{fig_dim_rd}. One can observe that the distributions of the original and generated series are highly overlapped, and the similarity is especially prominent for AQI and Traffic datasets (i.e., the 1st, 2nd, 4th, and 5th columns). These results visually validate that our generated datasets are indeed realistic across a variety of fields.

\textbf{Discriminative Score / MMD Score. } 
Other than visualization in low dimensional space, we further assess the generation quality, i.e., evaluate the similarity between the original and generated time-series, quantitatively using a neural-network-based discriminator and the MMD score. 
For the neural-network-based discriminator, by labeling the original time-series as positive samples and the generated time-series as negative ones, we train an LSTM classifier and then report the discriminative score in terms of $|\text{AUROC}-0.5|$ on the test set.
MMD is a frequently used metric to evaluate the similarity of two distributions \citep{grettonKernelMethodTwoSampleProblem2006}. It is estimated with $\widehat{\mathrm{MMD}}^2_u = \frac{1}{n(n-1)}\sum_{i=1}^n\sum_{j\neq i}^n K(x_i, x_j) - \frac{2}{mn} \sum_{i=1}^n \sum_{j=1}^m K(x_i, y_j) + \frac{1}{m(m-1)} \sum_{i=1}^m\sum_{j\neq i}^m K(y_i, y_j)$, where $K$ is the radial basis function (RBF) kernel. MMD gives another quantitative evaluation of the similarity without the need to train another neural network, as listed in the bottom row of Table \ref{tab_quat}. It is observed that the generated dataset is similar to the original ones, since the discriminative score is very close to zero (i.e., neural networks cannot distinguish generated samples from original samples), and the MMD score is relatively low. Other than discriminative score and MMD, we also utilize cross-correlation scores and perform additive experiments, which is shown in Supplementary Section \ref{app_results}.

\begin{table}[ht]
\centering
% \vspace{-2mm}
\setlength{\belowcaptionskip}{5pt}%
\caption{Quantitative assessment of the similarity between the generated and original time-series in terms of discriminative score and MMD. We show the ablation study in the table as well.}% The lower the score, the higher the similarity.}
\label{tab_quat}
\small
\renewcommand{\arraystretch}{1.05}
\renewcommand\tabcolsep{2pt}
\begin{tabular}{c|ccc|ccc}
 \toprule[.7pt]
\multirow{2}{*}{\textbf{Datasets}} & \multicolumn{3}{c|}{\textbf{Discriminative Score}}                  & \multicolumn{3}{c}{\textbf{MMD}}                                            \\
                           & AQI                     & Traffic                 & Medical                 & AQI                       & Traffic                   & Medical                 \\ \midrule[.4pt]
Additive Gaussian Noise    & 0.488 {\tiny $\pm$ 0.001} & 0.499 {\tiny $\pm$ 0.000} & 0.445 {\tiny $\pm$ 0.003} & 0.533 {\tiny $\pm$ 0.091}   & 0.716 {\tiny $\pm$ 0.011}   & 0.480 {\tiny $\pm$ 0.057} \\
w/o Noise Term             & 0.361 {\tiny $\pm$ 0.008} & 0.391 {\tiny $\pm$ 0.006} & 0.346 {\tiny $\pm$ 0.001} & 0.454 {\tiny $\pm$ 0.025}   & 0.717 {\tiny $\pm$ 0.007}   & \textbf{0.453 {\tiny $\pm$ 0.029}} \\
Fit w/o Residual Term      & 0.309 {\tiny $\pm$ 0.010} & 0.500 {\tiny $\pm$ 0.000} & 0.482 {\tiny $\pm$ 0.005} & 0.474 {\tiny $\pm$ 0.033}   & 0.858 {\tiny $\pm$ 0.020}   & 0.520 {\tiny $\pm$ 0.023} \\
Generate w/o Residual Term & 0.361 {\tiny $\pm$ 0.014} & 0.371 {\tiny $\pm$ 0.003} & 0.348 {\tiny $\pm$ 0.014} & 0.431 {\tiny $\pm$ 0.055}   & 0.657 {\tiny $\pm$ 0.016}   & 0.489 {\tiny $\pm$ 0.037} \\
Full Model                 & \textbf{0.054 {\tiny $\pm$ 0.025}} & \textbf{0.039 {\tiny $\pm$ 0.020}} & \textbf{0.017 {\tiny $\pm$ 0.027}} & \textbf{0.246 {\tiny $\pm$ 0.029}}   & \textbf{0.215 {\tiny $\pm$ 0.013}}   & \textbf{0.461 {\tiny $\pm$ 0.033}} \\ \bottomrule[.7pt]
\end{tabular}
\end{table}

\textbf{Ablation Study.~~~~} 
Using the above two quantitative scores, we also perform ablation studies to justify the effectiveness of our design. 
In the time-series fitting, we use normalizing flow to fit the noise distributions. To validate its effectiveness, we replace normalizing flow with \textbf{(a)} additive Gaussian noise (with parameters estimated from real series) and \textbf{(b)} no noise. 
Further, to validate that our pipeline reserves the real dynamics by splitting the causal model into causal term, residual term, and noise term, we add two alternatives that do not include the residual term when (c) fitting the NAR model or (d) generating new data, besides the above two settings.  
%\textbf{(c)} Fitting the time-series by only considering parents in HCG (where our approach is to fit with all time-series) and \textbf{(d)} Fit the time-series with all terms but generate new time-series with only causal term. 

The results are shown in Table \ref{tab_quat}, which shows that the full model produces time-series that mimic the original time-series best, in terms of both discriminative score and MMD score. The only exception is the slightly lower MMD under settings ``w/o Noise'' on medical datasets. It is worth noting that our discrimination scores are close to zero, i.e., neural networks cannot discriminate almost all generated time-series from original versions.

\subsection{Performance of State-of-the-art Causal Discovery Algorithms}
\label{exp_tscd}
To quantify the performances of different causal discovery algorithms, here we calculate their AUROC and AUPRC with respect to the ground truth causal graph. 
We do not evaluate the accuracy of the discovered causal graph $\mathbf{\tilde{A}}$ with respect to its ground-truth $\mathbf{\hat{A}}$, because there exists instantaneous effects in $\mathbf{x}_j \rightarrow \mathbf{x}_i^{\text{R}}$ (see Supplementary Section \ref{app_inst}). Instead, we ignore the blocks $\mathbf{I}_N, \mathbf{J}_N$ and $\mathbf{0}$ in Equation \ref{eqn_matrix}), and 
compare $\mathbf{\tilde{H}}$ with respect to $\mathbf{{H}}$, 
%only the $\mathbf{{H}}$ is compared with discovered $\mathbf{\tilde{H}}$ in terms of AUROC for causal discovery evaluation and The generated dataset $\left< \mathbf{\hat{X}}, \mathbf{\hat{A}} \right>$ is directly applicable for TSCD to get discovery result $\mathbf{\tilde{A}}$. However, 

% Moreover, the other three partitions of matrix $\mathbf{A}$ (i.e., $\mathbf{I}_N, \mathbf{J}_N, \mathbf{0}$ in Equation \ref{eqn_matrix}) is fixed and does not contain vital information. Consequently, only the $\mathbf{\tilde{H}}$ is the result we focus on.

\textbf{Baseline TSCD Algorithms.~~~~}   We benchmarked the performance of  9 most recent and representative causal discovery methods on our CausalTime datasets, including: 
% i) Granger-causality-based approach: 
Neural Granger Causality (NGC, \citep{tankNeuralGrangerCausality2022}); 
economy-SRU (eSRU, \citep{khannaEconomyStatisticalRecurrent2020}), a variant of SRU that is less prone to over-fitting when inferring Granger causality; 
Scalable Causal Graph Learning (SCGL, \citep{xuScalableCausalGraph2019}) that addresses scalable causal discovery problem with low-rank assumption; 
Temporal Causal Discovery Framework (TCDF, \citep{nautaCausalDiscoveryAttentionbased2019}) that utilizes attention-based convolutional neural networks; 
CUTS \citep{chengCUTSNeuralCausal2023} discovering causal relationships using two mutually boosting models, and CUTS+ \citep{chengCUTSHighdimensionalCausal2023} upgrading CUTS to high dimensional time-series. 
% ii) Constraint-based approaches:
constraint-based approach PCMCI \citep{rungeDetectingQuantifyingCausal2019}; 
% iii) CCM-based approaches:
Latent Convergent Cross Mapping (LCCM, \citep{brouwerLatentConvergentCross2021}); 
% iv) Other approaches:
Neural Graphical Model (NGM, \citep{bellotNeuralGraphicalModelling2022}), which employs neural ordinary differential equations to handle irregular time-series data.  
% {\color{red}shorten}
%We evaluated the performance in terms of the area under the ROC curve (AUROC) criterion.
To ensure fairness, we searched for the best set of hyperparameters for these baseline algorithms on the validation dataset, and tested performances on testing sets for 5 random seeds per experiment.

% \textbf{Existing Datasets.} As a reference, we also show testing performance on three frequently used datasets, two of which are numerical dataset, and another one is quasi dataset. i) Lorenz-96 dataset: simulated according to $\frac{dx_{i,t}}{dt} = -x_{i-1,t}(x_{i-2,t}-x_{i+1,t}) - x_{i,t} + F $, where we set $F=10, L=1000$. In this model, each time-series $\mathbf{x}_i$ is affected by historical values of four time-series $\mathbf{x}_{i-2}, \mathbf{x}_{i-1}, \mathbf{x}_{i}, \mathbf{x}_{i+1}$, and each row in the true causal graph $\mathbf{A}$ has four non-zero elements. ii) VAR dataset: simulated with the linear equation $\mathbf{x}_{:,t} = \sum_{\tau=1}^{\tau_{max}} \mathbf{A} \mathbf{x}_{:,t-\tau} + \mathbf{e}_{:,t} $, where the matrix $\mathbf{A}$ is the causal coefficients and $\mathbf{e}_{:,t}\sim \mathcal{N} (\mathbf{0},\sigma\mathbf{I})$. iii) Dream-3 dataset: a gene expression and regulation dataset widely used as causal discovery benchmarks \citep{khannaEconomyStatisticalRecurrent2020, tankNeuralGrangerCausality2022}. This dataset contains 5 models, each representing measurements of 100 gene expression levels. The length of each measured trajectory is $T=21$.

\begin{table}[ht]
% \vspace{-2mm}
\centering
\setlength{\belowcaptionskip}{5pt}%
\caption{Performance benchmarking of baseline TSCD algorithms on our CausalTime datasets. We highlight the best and the second best in bold and with underlining, respectively.}
\label{tab_tscd}
\small
\renewcommand{\arraystretch}{1.05}
\renewcommand\tabcolsep{3pt}
\begin{tabular}{c|ccc|ccc}
\toprule[.7pt]
\multirow{2}{*}{\textbf{Methods}}   & \multicolumn{3}{c|}{\textbf{AUROC}}                                                           & \multicolumn{3}{c}{\textbf{AUPRC}}                                                     \\ 
                        & AQI                                   & Traffic                               & Medical                               & AQI                                   & Traffic                               & Medical                        \\ \midrule[.4pt]
CUTS                    & 0.6013 {\tiny $\pm$ 0.0038}             & \textbf{0.6238 {\tiny $\pm$ 0.0179}}    & 0.3739 {\tiny $\pm$ 0.0297}             & 0.5096 {\tiny $\pm$ 0.0362}             & 0.1525 {\tiny $\pm$ 0.0226}                     & 0.1537 {\tiny $\pm$ 0.0039}      \\
CUTS+                   & \textbf{0.8928 {\tiny $\pm$ 0.0213}}    & \underline{0.6175 {\tiny $\pm$ 0.0752}} & \textbf{0.8202 {\tiny $\pm$ 0.0173}}    & \underline{0.7983 {\tiny $\pm$ 0.0875}} & \textbf{0.6367 {\tiny $\pm$ 0.1197}}            & 0.5481 {\tiny $\pm$ 0.1349}      \\
PCMCI                   & 0.5272 {\tiny $\pm$ 0.0744}             & 0.5422 {\tiny $\pm$ 0.0737}             & 0.6991 {\tiny $\pm$ 0.0111}             & 0.6734 {\tiny $\pm$ 0.0372}             & 0.3474 {\tiny $\pm$ 0.0581}                     & 0.5082 {\tiny $\pm$ 0.0177}      \\
NGC                     & 0.7172 {\tiny $\pm$ 0.0076}             & 0.6032 {\tiny $\pm$ 0.0056}             & 0.5744 {\tiny $\pm$ 0.0096}             & 0.7177 {\tiny $\pm$ 0.0069}             & 0.3583 {\tiny $\pm$ 0.0495}                     & 0.4637 {\tiny $\pm$ 0.0121}      \\
NGM                     & 0.6728 {\tiny $\pm$ 0.0164}             & 0.4660 {\tiny $\pm$ 0.0144}             & 0.5551 {\tiny $\pm$ 0.0154}             & 0.4786 {\tiny $\pm$ 0.0196}             & 0.2826 {\tiny $\pm$ 0.0098}                     & 0.4697 {\tiny $\pm$ 0.0166}      \\
LCCM                    & \underline{0.8565 {\tiny $\pm$ 0.0653}} & 0.5545 {\tiny $\pm$ 0.0254}             & \underline{0.8013 {\tiny $\pm$ 0.0218}} & \textbf{0.9260 {\tiny $\pm$ 0.0246}}    & \underline{0.5907 {\tiny $\pm$ 0.0475}}   & \textbf{0.7554 {\tiny $\pm$ 0.0235}}      \\
eSRU                    & 0.8229 {\tiny $\pm$ 0.0317}             & 0.5987 {\tiny $\pm$ 0.0192}             & 0.7559 {\tiny $\pm$ 0.0365}             & 0.7223 {\tiny $\pm$ 0.0317}             & 0.4886 {\tiny $\pm$ 0.0338}                     & \underline{0.7352 {\tiny $\pm$ 0.0600}}      \\
SCGL                    & 0.4915 {\tiny $\pm$ 0.0476}             & 0.5927 {\tiny $\pm$ 0.0553}             & 0.5019 {\tiny $\pm$ 0.0224}             & 0.3584 {\tiny $\pm$ 0.0281}             & 0.4544 {\tiny $\pm$ 0.0315}                     & 0.4833 {\tiny $\pm$ 0.0185}      \\
TCDF                    & 0.4148 {\tiny $\pm$ 0.0207}             & 0.5029 {\tiny $\pm$ 0.0041}             & 0.6329 {\tiny $\pm$ 0.0384}             & 0.6527 {\tiny $\pm$ 0.0087}             & 0.3637 {\tiny $\pm$ 0.0048}                     & 0.5544 {\tiny $\pm$ 0.0313}      \\  \bottomrule[.7pt]
\end{tabular}%\vspace{-1mm}
\end{table}

\textbf{Results and Analysis.~~~~} From the scores in Table \ref{tab_tscd}, one can see that among these algorithms, CUTS+ and LCCM perform the best, and most of the TSCD algorithms do not get AUROC $>0.9$. Interestingly, a few results demonstrate AUROC $< 0.5$, which means that we get inverted classifications.
% {\color{red} more comparison.}
The low accuracies tell that current TSCD algorithms still have a long way to go before being put into practice and indicate the necessity of designing more advanced algorithms with high feasibility to real data. 
Besides, compared with the reported results in previous work \citep{}, one can notice that the scores on our CausalTime dataset are significantly lower than those on synthetic datasets (e.g., VAR and Lorenz-96), on which some TSCD algorithms achieve scores close to 1. This implies that the existing synthetic datasets are insufficient to evaluate the algorithm performance on real data and calls for building new benchmarks to advance the development in this field.

\section{Additional Information}
We place theoretical analysis and assumptions in Supplementary Section \ref{app_theory}, implementation details along with hyperparameters for each of our key steps in Section \ref{app_details}, additional experimental results (including a comparison of various CDNN implementations, ablation study for scheduled sampling, and experimental results for cross-correlation scores) in Section \ref{app_results}, and algorithmic representation for our pipeline in Section \ref{app_algorithm}.

\section{Conclusions}

We propose CausalTime, a novel pipeline to generate realistic time-series with ground truth causal graphs, which can be used to assess the performance of TSCD algorithms in real scenarios and can also be generalizable to diverse fields. %Firstly, we use a neural network to identify the actual dynamics. Secondly, we divide the nonlinear autoregressive model into its causal component, residual element, and noise factor, and obtain the ground truth causal graph without losing its fidelity. Lastly, we create a new time-series with accurate causal graphs with the autoregression model. 
Our CausalTime contributes to the causal-discovery community by enabling upgraded algorithm evaluation under diverse realistic settings, which would advance both the design and applications of TSCD algorithms. Our code is available at \texttt{https://github.com/jarrycyx/unn}.

Our work can be further developed in multiple aspects. Firstly, replacing NAR with the SCM model can extend CausalTime to a richer set of causal models; secondly, we plan to take into account the multi-scale causal effects that widely exist in realistic time-series. 
Our future works include i) Creating a more realistic time-series by incorporating prior knowledge of certain dynamic processes or multi-scale associations. ii) Investigation of an augmented TSCD algorithm with reliable results on real time-series data.

\bibliography{ref}
\bibliographystyle{iclr2024_conference}

\newpage

\appendix
\section{Appendix}
\localtableofcontents

\secttoc

\subsection{Theory}
\label{app_theory}

\subsubsection{Assumptions}
\label{app_assumption}

Several assumptions are needed for our causal models, which are common in causal discovery literature.

\textbf{Markovian Condition.} The joint distribution can be factorized into $P(\mathbf{x}) = \prod_i P(x_i|\mathcal{P}(x_i))$, i.e., every variable is independent of all its nondescendants, conditional on its parents.

\textbf{Causal Faithfulness.} A causal model that accurately reflects the independence relations present in the data.

\textbf{Causal Sufficiency.} (or no latent confounder) All common causes of all variables are observed. This assumption is potentially very strong since we cannot observe ``all causes in the world'' because it may include infinite variables. However, this assumption is important for a variety of literature.

\textbf{No Instantaneous Effect.} (Or Temporal Priority \citep{assaadSurveyEvaluationCausal2022}) The cause occurred before its effect in the time-series. This assumption can be satisfied if the sampling frequency is higher than the causal effects. However, this assumption may be strong because the sampling frequency of many real time-series is not high enough.

\textbf{Causal Stationarity.} All the causal relationships remain constant throughout time. With this assumption, full time causal graph can be summarized into a windowed causal graph \citep{assaadSurveyEvaluationCausal2022}.

Other than these assumptions, we explicitly define a maximum time lag $\tau_{\text{max}}$ for causal effects (Section \ref{fitting}). However, long-term or multi-scale associations are common in many time-series \citep{dacuntoMultiscaleCausalStructure2022, wuTimesNetTemporal2DVariation2023}. We make this restriction because most causal discovery algorithm does not consider long-term or multi-scale associations.

\subsubsection{Instantaneous Effect of Residual Term}
\label{app_inst}

In section \ref{acg}, we propose to split the NAR model into causal term, residual term, and noise term, where the residual term $x_{t-1}^\text{R}$ is generated with
\begin{equation}
    \hat{x}^{\text{R}}_{t-1,i} = f_{\Theta_i} \left(\mathbf{\hat{x}}_{t-\tau:t-1,1}, ..., \mathbf{\hat{x}}_{t-\tau:t-1,N} \right)  - f_{\Theta_i} \left(\mathbf{\hat{x}}_{t-\tau:t-1,1}\cdot h_{1i}, ..., \mathbf{\hat{x}}_{t-\tau:t-1,N}\cdot h_{Ni} \right)
\end{equation}
As a result, instantaneous effects exist in this generation equation, i.e., in $\mathbf{x}_j \rightarrow \mathbf{x}_i^{\text{R}}$ but not in $\mathbf{x}_j \rightarrow \mathbf{x}_i$, $\forall i,j$. This is not a problem when tested on TSCD algorithms with compatibility with instantaneous effects. In the following, we discuss the consequences when tested on TSCD algorithms without compatibility to instantaneous effects, e.g., Granger Causality, Convergent Cross Mapping, and PCMCI \citep{rungeCausalInferenceTime2023}. For most cases, this does not affect causal discovery results between all $\mathbf{x}_j$. Though they may draw a wrong conclusion of $\mathbf{x}_j \nrightarrow \mathbf{x}_i^{\text{R}}$, this is not the part we compare to ground truth in the evaluation (Section \ref{exp_tscd}).

\textbf{Granger Causality.} Granger Causality determines causal relationships by testing if a time-series helps the prediction of another time-series. In this case, the causal discovery results of $\mathbf{x}_j \rightarrow \mathbf{x}_i$ are not affected since instantaneous effects of $\mathbf{x}_j \rightarrow \mathbf{x}_i^{\text{R}}$ do not affect whether the time-series $\mathbf{x}_j$ helps to predict the time-series $\mathbf{x}_i$, $\forall i, j$.

\begin{figure}[ht]
    \centering
    \includegraphics[width=0.8\linewidth]{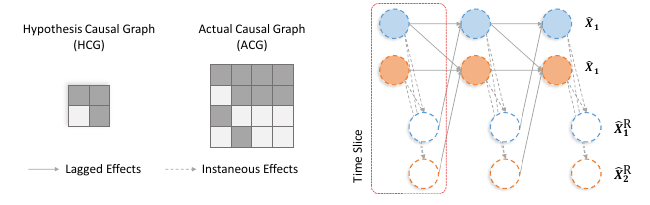}
    \caption{Illustration of instantaneous effects of $\mathbf{x}_j \rightarrow \mathbf{x}_i^{\text{R}}$. All paths from ${x}_{t_0,j}$ to ${x}_{t,i}$ through ${x}_{t,k}^{\text{R}}$ are blocked by conditioning on ${x}_{t,k}^{\text{R}}$, $\forall i,j,k$.}
    \label{app_fig_inst}
\end{figure}

\textbf{Constraint-based Causal Discovery.} These line of works is based on conditional independence tests, i.e., if $\mathbf{x}_{t_0,j} \rightarrow \mathbf{x}_{t,i}$, then ${x}_{t_0,j} \not\!\perp\!\!\!\perp {x}_{t,i} | \mathbf{X}_{t-\tau: t-1} \backslash \{{x}_{t_0,j}\}$, where $t-\tau \leq t_0 < t$ \citep{rungeDetectingQuantifyingCausal2019, rungeDiscoveringContemporaneousLagged2020}. This relationship is unaffected when considering instantaneous effect of ${x}_{t,j} \rightarrow {x}_{t,i}^{\text{R}}$, since all paths from ${x}_{t_0,j}$ to ${x}_{t,i}$ through ${x}_{t,k}^{\text{R}}$ are blocked by conditioning on ${x}_{t,k}^{\text{R}}$, $\forall i,j,k$, as is shown in Fig. \ref{app_fig_inst}.

\textbf{CCM-based Causal Discovery.} CCM detects if time-series $\mathbf{x}$ causes time-series $\mathbf{y}$ by examine whether time indices of nearby points on the $\mathbf{y}$ manifold can be used to identify nearby points on $\mathbf{x}$ \citep{sugiharaDetectingCausalityComplex2012, brouwerLatentConvergentCross2021}. In this case, the examination of whether $\mathbf{x}_j \rightarrow \mathbf{x}_i$ is not affected by $\mathbf{x}_k^{\text{R}}$, $\forall i,j,k$.

\subsection{Implementation Details}
\label{app_details}

\textbf{Network Structures and Training.}
The implementation of CDNN can vary and cMLP / cLSTM is not the only choice. For example, \citet{chengCUTSHighdimensionalCausal2023} explores enhancing causal discovery with a message-passing-based neural network, which is a special version of CDNN with extensive weight sharing. However, the fitting accuracy of CDNN is less explored. Sharing partial weights may alleviate the structural redundancy problem. Moreover, the performance with more recent structures is unknown, e.g. Transformer. In the following, this work investigates various implementations of CDNN when they are applied to fit causal models.

\begin{figure}[ht]
    \centering
    \includegraphics[width=\linewidth]{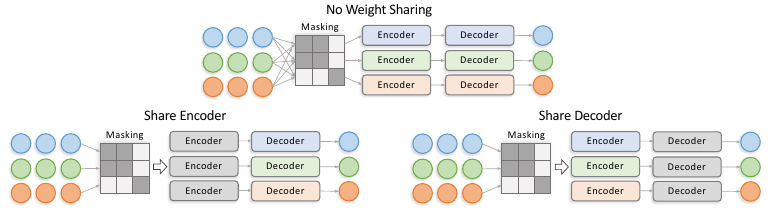}
    \vspace{-3mm}
    \caption{Illustration of network structures of CDNN.}
    \label{fig_net}
\end{figure}

Specifically, three kinds of backbones combining three network sharing policies are applied, i.e., MLP, LSTM, Transformer combining no sharing, shared encoder, and shared decoder. For MLP, the encoder and decoder are both MLP. For LSTM, we assign an LSTM encoder and an MLP decoder. For the Transformer, we assign a Transformer encoder and MLP decoder. We show the structure for no sharing, shared encoder, and shared decoder in Figure \ref{fig_net}, with a three-variable example. The test results for prediction using these architectures are shown in Table \ref{app_tab_net}.

\begin{table}[ht]
\setlength{\belowcaptionskip}{5pt}%
\caption{Hyper parameters for time-series fitting.}
\label{app_tab_fit_param}
\renewcommand{\arraystretch}{1.2}
\centering
\small
\setlength{\tabcolsep}{15pt}
\label{app_param}
\begin{tabular}{c|c|ccccc}
\toprule[.7pt]
Module                              & Parameter     & AQI & Traffic & Medical \\ \midrule[.4pt]
\multirow{3}{*}{LSTM Encoder}       & Layers        &  2   &   2      &   2      \\
                                    & Hidden        &  128 & 128  &128    \\
                                    & Heads        & 4 &   4   & 4   \\ \midrule[.4pt]
\multirow{2}{*}{MLP Encoder}        & Layers        & 3   &    3    &      3  \\
                                    & Hidden        & 128&    128 &   128  \\ \midrule[.4pt]
\multirow{2}{*}{Transformer Encoder}& Hidden        & 128 &  128    &  128   \\
                                    & Heads         &  4  &   4     &  4    \\ \midrule[.4pt]
\multirow{2}{*}{Decoder}            & Layers        &  3  &    3    &  3      \\
                                    & Hidden        & 128 &   128   & 128     \\ \midrule[.4pt]
\multirow{3}{*}{Training}           & Learning Rate &  0.01   & 0.001 &   0.003      \\
                                    & Optimizer     &   Adam&  Adam   & Adam   \\
                                    & Input Window  &   20  &  20     & 20     \\ \midrule[.4pt]
\multirow{2}{*}{Normalizing Flow}   & Layers        &  5  &    5   &     5   \\
                                    & Hidden        &  128  &    64   & 64    \\
\bottomrule[.7pt]
\end{tabular}
\end{table}

\begin{table}[ht]
\setlength{\belowcaptionskip}{5pt}%
\setlength{\tabcolsep}{15pt}
\caption{Hyperparameters settings of the baseline causal discovery and data imputation algorithms.}
\label{param}
\centering
\label{baseline_hyper}
\footnotesize
\begin{tabular}{c|c|ccc}
\toprule[.7pt]
\textbf{Methods}       & \textbf{Params.}           & \textbf{AQI}                 & \textbf{Traffic}   & \textbf{Medical}   \\ \midrule[.4pt]
\multirow{2}{*}{PCMCI} & $\tau_{max}$               & 5                            & 5                  & 5                  \\
                       & $PC_{\alpha}$              & 0.05                            & 0.05                  & 0.05               \\ \midrule[.4pt]
\multirow{3}{*}{NGC}   & Learning rate              & 0.05                         & 0.05               & 0.05               \\
                       & $\lambda_{ridge}$          & 0.01                         & 0.01               & 0.01               \\
                       & $\lambda$                  & $0.02\rightarrow 0.2$        & $0.02\rightarrow 0.2$ & $0.02\rightarrow 0.2$  \\ \midrule[.4pt]
\multirow{4}{*}{eSRU}  & $\mu_1$                    & 0.1                          & 0.1                & 0.7                \\ 
                       & Learning rate              & 0.01                         & 0.01               & 0.001              \\ 
                       & Batch size                 & 40                          & 40                & 40                \\ 
                       & Epochs                     & 50                         & 50               & 50               \\ \midrule[.4pt]
\multirow{3}{*}{SCGL}  & $\tau$                    & 10                           & 10                 & 10                 \\
                       & Batch size                 & 32                           & 32                 & 32                 \\
                       & Window                     & 3                            & 3                  & 3                  \\ \midrule[.4pt]
\multirow{3}{*}{LCCM}  & Epochs                     & 50                           & 50                 & 50                 \\
                       & Batch size                 & 10                           & 10                 & 10                 \\
                       & Hidden size                & 20                           & 20                 & 20                 \\ \midrule[.4pt]
\multirow{3}{*}{NGM}   & Steps                      & 200                         & 200               & 200               \\
                       & Horizon                    & 5                            & 5                  & 5                  \\
                       & GL\_reg                    & 0.05                         & 0.05               & 0.05               \\ \midrule[.4pt]
\multirow{2}{*}{TCDF}  &$\tau $                  & 10                            & 10                  & 10                  \\
                       &Epoch num                   & 1000                         & 1000               & 1000               \\ 
                       &Learning rate               & 0.01                         & 0.01               & 0.01                  \\ \midrule[.4pt]
\multirow{3}{*}{CUTS}  &Input step                  & 20                            & 20                  & 20                  \\
                       &$\lambda$                   & 0.1                          & 0.1                & 0.1\\
                       &       $\tau $   & $0.1\rightarrow 1$&$0.1\rightarrow 1$ &$0.1\rightarrow 1$\\ \midrule[.4pt]
\multirow{3}{*}{CUTS+} &Input step                  & 1                            & 1                  & 1                  \\
                       &$\lambda$                   & 0.01                          & 0.01                & 0.01\\
                       &       $\tau $   & $0.1\rightarrow 1$&$0.1\rightarrow 1$ &$0.1\rightarrow 1$\\ \bottomrule[.7pt]
\end{tabular}

\end{table}

\textbf{Normalizing Flow.} We implement normalizing flow using its open-source repository\footnote{https://github.com/VincentStimper/normalizing-flows}. We use normal distributions as base distributions. For transformations, we use simple combinations of linear and nonlinear layers, with parameters shown in Table \ref{app_tab_fit_param}.

\textbf{DeepSHAP.} We use the official implementation of DeepSHAP\footnote{https://github.com/shap/shap}. Specifically, we use its ``DeepExplainer'' module to explain the time-series prediction model which is trained with a fully connected graph. Note that this prediction model is a CDNN, which enables a pairwise explanation of feature $i$'s importance to the prediction of $j$. The explained samples are randomly selected from a train set of real time-series, and the final feature importance graph $\Phi$ is acquired by taking the average values of all samples. To convert to binarized HCG, we select a threshold to get a sparsity of 15\% (i.e., 15\% of the elements in HCG are labeled as 1).

\textbf{Prior Graph Extraction.} For datasets with prior knowledge, e.g., AQI and Traffic, relationships between each variable are highly relevant to geometry distances. Consequently, we extract HCG from the geographic distances between nodes using a thresholded Gaussian kernel, i.e., 
\begin{equation}
w^{ij} = 
\begin{cases}
1, & \text{dist}(i, j) \leq \sigma \\
0, & \text{otherwise}
\end{cases}
\end{equation}
we select $\sigma$ based on geographical distances (which is $\approx 40$ km for AQI dataset and for Traffic dataset, we use the ``dist\_graph'' from \url{https://github.com/liyaguang/DCRNN/tree/master}).

\textbf{Dimension Reduction.} For the implementation of t-SNE and PCA, we use scikit-learn\footnote{https://scikit-learn.org/} package. To solve the dimension reduction in an acceptable time, we split the generated time-series into short sequences (with lengths of 5) and flattened them for the input of dimension reduction.

\textbf{Autoregressive Generation.} After fitting the time-series with neural networks and normalizing flow, and acquiring the ground-truth causal graph by splitting the causal model, we generate a new time-series autoregressively. Although we utilize scheduled sampling to avoid the accumulation of the generation error, the total time step must be limited to a relatively small one. Actually, our generated time length is 40 for these three datasets. For each of them, we generate 500 samples, i.e., a total of 20000 time steps for each dataset.

\textbf{TSCD Algorithm Evaluation.} Since our generated time-series are relatively short and contain several samples, we alter existing approaches by enabling TSCD from multiple observations (i.e. multiple time-series). For neural-network-based or optimization-based approaches such as CUTS, CUTS+, NGC, and TCDF, we alter their dataloader module to prevent cross-sample data fetching. For PCMCI, we use its variant JPCMCI+ which permits the input of multiple time-series. For remaining TSCD algorithms that do not support multiple time-series, we use zero-padding to isolate each sample. We list the original implementations of our included TSCD algorithms in the following:
\begin{itemize}[leftmargin=*]
    \item \textit{PCMCI.~~~~} The code is from \texttt{https://github.com/jakobrunge/tigramite}.

    \item \textit{NGC.~~~~} The code is from \texttt{https://github.com/iancovert/Neural-GC}. We use the cMLP network because according to the original paper \citep{tankNeuralGrangerCausality2022} cMLP achieves better performance, except for Dream-3 dataset.

    \item \textit{eSRU.~~~~} The code is from \texttt{https://github.com/sakhanna/SRU\_for\_GCI}.

    \item \textit{SCGL.~~~~} The code is downloaded from the link shared in its original paper \citep{xuScalableCausalGraph2019}.

    \item \textit{LCCM.~~~~} The code is from \texttt{https://github.com/edebrouwer/latentCCM}.

    \item \textit{NGM.~~~~} The code is from \texttt{https://github.com/alexisbellot/Graphical-modelling- continuous-time}.

    \item \textit{CUTS / CUTS+.~~~~} The code is from \texttt{https://github.com/jarrycyx/UNN}.

    \item \textit{TCDF.~~~~} The code is from \texttt{https://github.com/M-Nauta/TCDF}.
\end{itemize}

\textbf{Discriminative Network.} To implement the discrimination score in time-series quality control, we train separate neural networks for each dataset to classify the original from the generated time-series. We use a 2-layered LSTM a with hidden size of 8, the training is performed with a learning rate of 1e-4 and a total of 30 epochs.

\subsection{Additional Results}
\label{app_results}

\subsubsection{Time-series Fitting}

In Section \ref{fitting}, we show that CDNN can be used to fit causal models with adjacency matrix $\mathbf{A}$. By splitting each network into two parts, i.e., encoder and decoder, 9 combinations (MLP, LSTM, Transformer combining shared encoder, shared decoder, and no weight sharing) are considered in the experiments. By comparing fitting accuracy (or prediction accuracy) on the AQI dataset, we observe that LSTM with a shared decoder performs the best among 9 implementations. 

\begin{table}[]
\setlength{\belowcaptionskip}{5pt}%
\setlength{\tabcolsep}{9pt}
\caption{Comparison of predictive MSE with different implementations of CDNN. The experiments are performed on AQI datasets.}
\label{app_tab_net}
\small
\centering
\begin{tabular}{cc|ccc}
\toprule[.7pt]
\multicolumn{2}{c|}{\textbf{Backbone}}                    & \textbf{MLP}             & \textbf{LSTM}            & \textbf{Transformer}     \\ \midrule[.4pt]
\multirow{3}{*}{Weight Sharing} & No Sharing     & $0.0511\pm 0.0009$ & $0.0017\pm 0.0002$ & $.0061\pm 0002$ \\
                                & Shared Encoder & $0.0248\pm 0.0033$ & $0.0048\pm 0.0002$ & $0.0056\pm 0.0001$ \\
                                & Shared Decoder & $0.0254\pm 0.0024$ & \textbf{0.0014 $\pm$ 0.0002} & $0.0060\pm 0.0002$ \\ \bottomrule[.7pt]
\end{tabular}
\end{table}

\subsubsection{Scheduled Sampling}

To validate if scheduled sampling is effective in the training process, we perform an ablation study on three datasets with different autoregressive prediction steps. We observe in Table \ref{app_tab_stab} that, by incorporating scheduled sampling, the cumulative error decreases, and the decrement is large with higher autoregressive steps. This demonstrates that scheduled sampling does decrease accumulative error for the fitting model, which is beneficial for the following generation process.

\begin{table}[ht]
\setlength{\belowcaptionskip}{5pt}%
\setlength{\tabcolsep}{10pt}
\caption{Comparison of prediction MSE with and without scheduled sampling policy. The comparisons are performed with different autoregressive prediction steps $n$ in terms of MSE.}
\label{app_tab_stab}
\small
\centering
\begin{tabular}{c|cc|cc|cc}
\toprule[.7pt]
\multirow{2}{*}{\textbf{Step}} & \multicolumn{2}{c|}{\textbf{AQI}} & \multicolumn{2}{c|}{\textbf{Traffic}} & \multicolumn{2}{c}{\textbf{Medical}} \\
                      & w          & w/o        & w            & w/o          & w            & w/o          \\ \midrule[.4pt]
$n=2$                   & 0.0129     & 0.0123     & 0.0124       & 0.0124       & 0.0082       & 0.0082       \\
$n=5$                   & 0.0348     & 0.0329     & 0.0279       & 0.0282       & 0.0136       & 0.0155       \\
$n=10$                  & 0.0351     & 0.0377     & 0.0312       & 0.0354       & 0.0181       & 0.0193       \\
$n=20$                  & 0.0331     & 0.034      & 0.0318       & 0.0331       & 0.0144       & 0.0174       \\ \bottomrule[.7pt]
\end{tabular}
\end{table}

\subsubsection{Cross Correlation Scores for Time-series Generation}

Despite the discriminative score and MMD score, we further compare the similarity of generated data to original versions in terms of cross-correlation scores. Specifically, we calculate the correlation between real and generated feature vectors, and then report the sum of the absolute differences between them, which is similar to \citet{jarrettTimeseriesGenerationContrastive2021}'s calculation process. We show the results along with the ablation study in Table \ref{app_tab_corr}.

\begin{table}[ht]
\centering
\setlength{\belowcaptionskip}{5pt}%
\caption{Quantitative assessment of the similarity between the generated and original time-series in terms of cross-correlation scores. We show the ablation study in the table as well.}% The lower the score, the higher the similarity.}
\label{app_tab_corr}
\small
\renewcommand{\arraystretch}{1.05}
\renewcommand\tabcolsep{10pt}
\begin{tabular}{c|ccc}
 \toprule[.7pt]
\multirow{2}{*}{\textbf{Datasets}} & \multicolumn{3}{c}{\textbf{Cross Correlation Score}}                                               \\
                           & AQI                                & Traffic                           & Medical                           \\ \midrule[.4pt]
Additive Gaussian Noise    & 43.74 {\tiny $\pm$ 11.55}          & 198.30 {\tiny $\pm$ 4.00}         & 18.91 {\tiny $\pm$ 2.17}          \\
w/o Noise Term             & 50.04 {\tiny $\pm$ 4.79}           & 194.44 {\tiny $\pm$ 5.02}         & \textbf{20.77 {\tiny $\pm$ 4.40}} \\
Fit w/o Residual Term      & 49.04 {\tiny $\pm$ 7.01}           & 238.38 {\tiny $\pm$ 4.90}         & \textbf{21.94 {\tiny $\pm$ 3.45}} \\
Generate w/o Residual Term & 40.62 {\tiny $\pm$ 24.73}          & 164.14 {\tiny $\pm$ 3.04}         & 23.53 {\tiny $\pm$ 3.12}          \\
Full Model                 & \textbf{39.75 {\tiny $\pm$ 5.24}}  & \textbf{60.37 {\tiny $\pm$ 2.88}} & \textbf{22.37 {\tiny $\pm$ 1.59}} \\ \bottomrule[.7pt]
\end{tabular}
\end{table}

\subsection{Algorithmic Representation for CausalTime Pipeline}

We show the detailed algorithmic representation of our proposed data generation pipeline in Algorithm \ref{app_algo_ppl}, where we exclude quality control and TSCD evaluation steps.

\begin{algorithm}[ht]
\label{app_algo_ppl}
\setstretch{1.15}
    \renewcommand{\algorithmicrequire}{\textbf{Input:}}
	\renewcommand{\algorithmicensure}{\textbf{Output:}}
	\caption{Pipeline for CausalTime Generation (Excluding quality control and TSCD evaluation)} 
	\label{app_algorithm} 
	\begin{algorithmic}
		\REQUIRE Time-series dataset $\mathbf{X} = \left\{ \vx_{1:L, 1}, ... ,\vx_{1:L, N} \right\}$ with length $L$; (Optional) graph $\mathbf{H}_p$ generated with prior knowledge; generation length $\hat{L}$.
		\ENSURE Paired time-series causal discovery (TSCD) dataset $\left< \mathbf{\hat{X}}, \mathbf{\hat{A}} \right>$.
            \STATE Initilize parameter set $\{\Theta_i\}_{i=1}^N$, $\{\Psi_i\}_{i=1}^N$

		\STATE \textcolor{blue}{\texttt{\# Time-series Fitting}}
		\FOR {$n_1$ Epochs}
                \STATE Update $\{\Theta_i\}_{i=1}^N$ with Algorithm \ref{app_pred} (Prediction Model Fitting).
		\ENDFOR
		\FOR {$n_1$ Epochs}
                \STATE Update $\{\Psi_i\}_{i=1}^N$ with Algorithm \ref{app_nf} (Noise Distribution Fitting).
		\ENDFOR

		\STATE \textcolor{blue}{\texttt{\# HCG Extraction}}
            \IF {Exists prior knowledge $\mathbf{H}_p$}
                \STATE $\mathbf{H} \leftarrow \mathbf{H}_p$
            \ELSE
                \STATE $\mathbf{H} \leftarrow \left\{ \text{DeepSHAP}\left( {f_{\Theta_i}}(\cdot) \right) \right\}_{i=1}^N$
            \ENDIF
            \STATE $\mathbf{H} \leftarrow \mathbb{I}(\mathbf{H} > \gamma)$, where calculation of $\gamma$ is discussed in Section \ref{hcg}. 
            
		\STATE \textcolor{blue}{\texttt{\# Time-series Generation with ACG}}
            \STATE Generate actual causal graph $\mathbf{\hat{A}}$ with Equation \ref{eqn_matrix}, select initial sequence $\mathbf{\hat{X}}_{0:\tau-1}$ randomly from original time-series $\mathbf{X}$.
            \FOR {$t = \tau$ top $\hat{L}$}
                \FOR {$i = 1$ top $N$}
                    \STATE Generate $\hat{x}_{t,i}$ and $\hat{x}_{t,i}^{\text{R}}$ with Equation \ref{eqn_split} and \ref{eqn_res}, where $\hat{\eta}_{t,i}$ is sampled from $T_{\Psi_i}(u)$.
                \ENDFOR
            \ENDFOR
		\RETURN Generated TSCD dataset $\left< \mathbf{\hat{X}}, \mathbf{\hat{A}} \right>$.
	\end{algorithmic} 
\end{algorithm}

\begin{algorithm}[ht]
\setstretch{1.15}
    \renewcommand{\algorithmicrequire}{\textbf{Input:}}
	\renewcommand{\algorithmicensure}{\textbf{Output:}}
	\caption{Prediction Model Fitting} 
	\label{app_pred} 
	\begin{algorithmic}
		\REQUIRE Time-series dataset $\left\{ \vx_{1:L, 1}, ... ,\vx_{1:L, N} \right\}$ with length $L$; parameter set $\{\Theta_i\}_{i=1}^N$.
		\ENSURE Paired time-series causal discovery (TSCD) dataset $\left< \mathbf{\hat{X}}, \mathbf{\hat{A}} \right>$.

              \FOR {$i = 1$ top $N$}
                    \STATE Perform prediction with $\hat{x}_{t,i} \leftarrow f_{\Theta_i} \left(\mathbf{x}_{t-\tau:t-1,1}, ..., \mathbf{x}_{t-\tau:t-1,N} \right) + \eta_{t,i}$
                    \STATE Calculate loss function with $\text{MSE}(\hat{X}, {X})$
                \STATE Update $\{\Theta_i\}_{i=1}^N$ with Adam optimizer
              \ENDFOR
		\RETURN Discovered causal adjacency matrix $\hat{\mA}$ where each elements is $\tilde{a}_{i,j}$.
	\end{algorithmic} 
\end{algorithm}

\begin{algorithm}[ht]
\setstretch{1.15}
    \renewcommand{\algorithmicrequire}{\textbf{Input:}}
	\renewcommand{\algorithmicensure}{\textbf{Output:}}
	\caption{Noise Distribution Fitting} 
	\label{app_nf} 
	\begin{algorithmic}
		\REQUIRE Time-series dataset $\left\{ \vx_{1:L, 1}, ... ,\vx_{1:L, N} \right\}$ with length $L$; Parameter set $\{\Theta_i\}_{i=1}^N$, $\{\Psi_i\}_{i=1}^N$.
		\ENSURE Paired time-series causal discovery (TSCD) dataset $\left< \mathbf{\hat{X}}, \mathbf{\hat{A}} \right>$.

              \FOR {$i = 1$ top $N$}
                    \STATE Perform prediction with $\hat{x}_{t,i} \leftarrow f_{\Theta_i} \left(\mathbf{x}_{t-\tau:t-1,1}, ..., \mathbf{x}_{t-\tau:t-1,N} \right) + \eta_{t,i}$.
                    \STATE Calculate real noise $\eta_{t,i} \leftarrow x_{t,i} - \hat{x}_{t,i}$.
                    \STATE Calculate likelihood $p_{\eta_i}(\eta_{t,i}) \leftarrow p_u(T^{-1}_{\psi_i}(\eta_{t,i})) \frac{\partial}{\partial \eta_{t,i}} T^{-1}_{\psi_i}(\eta_{t,i})$.
                    \STATE Update $\{\Psi_i\}_{i=1}^N$ with Adam optimizer.
              \ENDFOR
		\RETURN Discovered causal adjacency matrix $\hat{\mA}$ where each elements is $\tilde{a}_{i,j}$.
	\end{algorithmic} 
\end{algorithm}

\end{document}

%% file: math_commands.tex
\usepackage{amsmath,amsfonts,bm}

\def\eqref#1{equation~\ref{#1}}
\def\1{\bm{1}}

\def\vx{{\bm{x}}}

\def\mA{{\bm{A}}}

\def\mX{{\bm{X}}}

\DeclareMathAlphabet{\mathsfit}{\encodingdefault}{\sfdefault}{m}{sl}
\SetMathAlphabet{\mathsfit}{bold}{\encodingdefault}{\sfdefault}{bx}{n}